\documentclass[numsec,webpdf,modern,large]{oup-authoring-template}

\graphicspath{{Fig/}}
\newcommand{\societylogo}{}

\usepackage{amsmath}
\usepackage{amssymb}
\usepackage{dsfont}
\usepackage{csquotes}
\usepackage{xspace}
\usepackage{booktabs}
\usepackage{pifont}

\definecolor{myred}{RGB}{220, 50, 32}
\definecolor{myblue}{RGB}{0, 90, 181}

\newcommand{\tick}{\ding{51}}
\newcommand{\cross}{\ding{55}}

\newcommand{\gene}{{\normalsize G\scriptsize ENE}\xspace}
\newcommand{\disease}{{\normalsize D\scriptsize ISEASE}\xspace}
\newcommand{\chemical}{{\normalsize C\scriptsize HEMICAL}\xspace}
\newcommand{\species}{{\normalsize S\scriptsize PECIES}\xspace}
\newcommand{\cellline}{{\normalsize C\scriptsize ELL \normalsize L\scriptsize INE}\xspace}
\newcommand{\variant}{{\normalsize V\scriptsize ARIANT}\xspace}

\newcommand{\ncbigene}{\textsf{NCBI Gene}\xspace}
\newcommand{\ncbitaxonomy}{\textsf{NCBI Taxonomy}\xspace}
\newcommand{\cellosaurus}{\textsf{Cellosaurus}\xspace}
\newcommand{\dbsnp}{\textsf{dbSNP}\xspace}
\newcommand{\mesh}{\textsf{MeSH}\xspace}
\newcommand{\umls}{\textsf{UMLS}\xspace}

\newcommand{\ctddiseases}{\textsf{CTD Diseases}\xspace}
\newcommand{\ctdchemicals}{\textsf{CTD Chemicals}\xspace}

\newcommand{\nlmgene}{\texttt{NLM-Gene}\xspace}
\newcommand{\gnormplus}{\texttt{GNormPlus}\xspace}
\newcommand{\linnaeus}{\texttt{Linnaeus}\xspace}
\newcommand{\esseight}{\texttt{S800}\xspace}

\newcommand{\bioid}{\texttt{BioID}\xspace}

\newcommand{\osiris}{\texttt{OSIRIS v1.2}\xspace}
\newcommand{\snp}{\texttt{SNP}\xspace}
\newcommand{\tmvarthree}{\texttt{tmVar v3}\xspace}
\newcommand{\ncbidisease}{\texttt{NCBI Disease}\xspace}
\newcommand{\bcfivecdr}{\texttt{BC5CDR}\xspace}

\newcommand{\biored}{\texttt{BioRED}\xspace}

\newcommand{\nlmchem}{\texttt{NLM-Chem}\xspace}

\newcommand{\medmentionspv}{\texttt{MedMentions21}\xspace}

\newcommand{\belxtr}{BELXTR\xspace}

\begin{document}

\journaltitle{Journal Title Here}
\DOI{DOI added during production}
\copyrightyear{YEAR}
\pubyear{YEAR}
\vol{XX}
\issue{x}
\access{Published: Date added during production}
\appnotes{Paper}

\firstpage{1}


\title[BELXTR]{BELXTR: Biomedical Entity Linking via Contextualized Token Retrieval}

\author[1,$\ast$]{Samuele Garda}
\author[1]{Ulf Leser}

\address[1]{\orgdiv{Computer Science}, \orgname{Humboldt-Universit\"at zu Berlin}, \orgaddress{\street{Rudower Chaussee 25}, \postcode{12489}, \state{Berlin}, \country{Germany}}}

\corresp[$\ast$]{Corresponding author. \href{gardasam@informatik.hu-berlin.de}{gardasam@informatik.hu-berlin.de}}

\received{Date}{0}{Year}
\revised{Date}{0}{Year}
\accepted{Date}{0}{Year}




\abstract{
	\textbf{Motivation}:  Biomedical Entity Linking disambiguates mentions to entities in a knowledge base (KB),
	making it the cornerstone of information extraction pipelines. While embedding-based models are a popular approach for the task,
	they suffer from a key limitation. They compress mentions (and entities) into a single vector, forcing the model to average away crucial
	fine-grained differences. \\
	\textbf{Results}: We present BELXTR, a novel embedding model based on the multi-vector (a.k.a. late interaction) architecture, which
	allows to leverage token-level matching information.
	BELXTR extends the original XTR model to biomedical entity linking
	by integrating an existing task-specific training objective and exploring active query expansion.
	Experiments across ten corpora and five KBs show that BELXTR improves upon current state-of-the-art in half
	of the corpora with an average improvement of 5pp recall@1. The largest gains are reported on the challenging
	cross-species gene disambiguation subtask, where BELXTR outperforms an LLM-powered retrieve-and-rerank pipeline and
	closely approaches a specialized rule-based system.
	Our results highlight multi-vector models as a practical alternative to hard-to-maintain rule-based systems or in scenarios where LLM-based
	reranking is too costly as in PubMed-scale mining. \\
	\textit{Availability and implementation}: The code to reproduce our experiments can be found at:
	\href{https://github.com/sg-wbi/belxtr}{https://github.com/sg-wbi/belxtr}.\\
}

%


%
%
\maketitle



\section{Introduction}

Biomedical Entity Linking\footnote{Also known as Named Entity Normalization or Entity Disambiguation. We will use
	\enquote{linking}, \enquote{normalization}, and \enquote{disambiguation} interchangeably throughout the text.} (BEL) is
the task of disambiguating mentions of biomedical concepts to unique entries in knowledge base\footnote{We will use the
	term to refer to any type of normalization target, including ontologies and controlled vocabularies.} (KB). As a key
component in the conversion from text to structured representation, BEL supports multiple downstream applications such
as information retrieval \citep{10.1007/978-3-032-04354-2_12} and knowledge graph construction \citep{SCHAFER2024639}.

Embedding-based models are one of the most popular approaches for the task
\citep{SelfAlignmentLiuF2021,biomedicalentisung2020,bern2anadvanmujeen}. However, they suffer from a fundamental
representational bottleneck. As they compress mentions (and entity names) into a single vector, they are forced to
average away fine-grained surface-form variations, which are crucial for correct disambiguation.

Multi-vector (a.k.a late-interaction\footnote{We use the terms \enquote{late-interaction} and \enquote{multi-vector}
	interchangeably throughout the text.}) models are a promising approach to overcome this limitation. They generate
individual subword embeddings for every token in the mention and entity, and use the pooled Cartesian product of their
similarities to produce the final similarity score. However, multi-vector models have not been thoroughly investigated
for entity linking, particularly in specialized fields like biomedicine, as existing work remains almost exclusively
confined to applying a vanilla ColBERT \citep{khattab2020colbert} model to general-domain datasets
\citep{Understanding_H_Zhang_2021,song-etal-2024-comparing}.

Here we introduce \belxtr, a multi-vector model specifically developed for biomedical entity linking. \belxtr extends
the original XTR model \citep{lee2023rethinking} (an improved version of the ColBERT model \citep{khattab2020colbert})
by (i) using an existing BEL-specific training objective and (ii) optimizing temperature scaling for contrastive
learning (as proposed by \citep{Learning_transf_Radfor_2021}). We explore as well the explicit training of
\texttt{[MASK]} embeddings to perform query\footnote{A contextualized mention.} expansion via an auxiliary objective
function (see Section \ref{sec:xtr:method:model:qe} for details).

We evaluate \belxtr on the standardized BELB benchmark \citep{belb}, comparing it against six state-of-the-art neural
models. \belxtr outperforms all baselines on five of the ten evaluated corpora and achieves the second-best performance
on four, yielding an average improvement of 5 percentage points in recall@1 over the best performing models. Results on
the \gene corpora drive this performance increase the most. We argue that \belxtr's edge stems from its ability to
model fine-grained, contextualized token similarities. For instance, as shown in Figure~\ref{fig:late-interaction}, the
model can learn to directly capture subtle differences such as \enquote{$\alpha$2-} vs.
\enquote{$\beta$2-microglobulin} (see Section \ref{sec:xtr:dicussion:gene} for further discussion).

Notably, on the challenging cross-species gene disambiguation task, \belxtr as a standalone retriever achieves a higher
recall@1 than a LLM-powered retrieve-and-rerank pipeline (see Section \ref{sec:xtr:discussion:llm} for discussion). We
compare \belxtr as well against the state-of-the-art, traditional type-specific systems integrated in PubTator3
\citep{pubtator3}. While maintaining competitive results across five entity types, on gene disambiguation our model
achieves an F1 score of 83.86, outperforming the previous best neural model (80.50) and closely approaching PubTator3's
highly specialized \gene model (84.63).

Overall, in line with findings in information retrieval studies \citep{thakur2021beir,warner2024smarter}, our results
show that multi-vector models yield superior retrieval capabilities, advancing the state-of-the-art on the BEL task.
This makes them a practical alternative to hard-to-maintain rule-based systems or in scenarios where LLM-based
reranking is too costly (e.g., PubMed-scale mining).

\section{Materials and methods}~\label{sec:xtr:method}

We now introduce (i) \belxtr, our novel method for biomedical entity linking (Section \ref{sec:xtr:method:model}) based
on the XTR model first introduced by \cite{lee2023rethinking} and (ii) the evaluation protocols adopted in our
experiments (Section \ref{sec:xtr:method:eval}).

\subsection{Model}\label{sec:xtr:method:model}

\begin{figure*}[!htbp]
	\centering
	\includegraphics[scale=0.8]{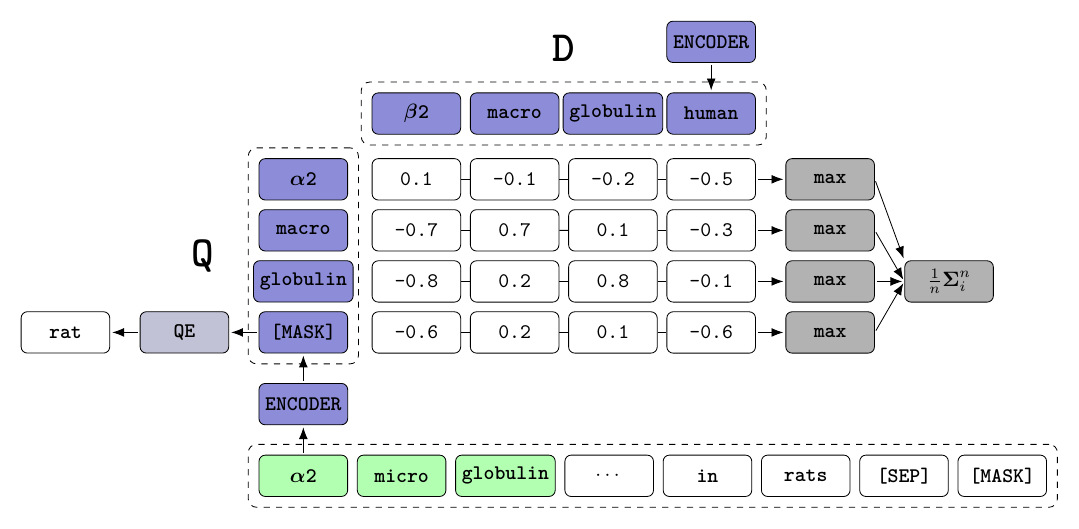}
	\caption{Illustration of BELXTR: a multi-vector model for biomedical entity linking.
		Green background indicates mention boundaries marked by special tokens (omitted for clarity).
		Q: mentions with context. D: entity name.
		QE: Query Expansion (see Section \ref{sec:xtr:method:model:qe}).
	}
	\label{fig:late-interaction}
\end{figure*}

\subsubsection{Background}%
Before introducing our enhancements (see Section \ref{sec:xtr:results:ablation} for the ablation study), we first
revisit the late-interaction model on which our method is based, namely XTR \citep{lee2023rethinking}.

Given a mention (with its context) and a candidate entity,\footnote{While late-interaction models are traditionally
	utilized for information retrieval \citep{DBLP:conf/cikm/ChaffinS25}, we describe them here in the context of entity
	linking.} XTR maps them to token embedding matrices $\boldsymbol{Q} \in \mathbb{R}^{m \times h}$ and $\boldsymbol{D}
	\in \mathbb{R}^{n \times h}$, respectively. Here $n$, $m$, and $h$ denote the number of mention tokens, the number of
entity tokens, and the embedding dimension, respectively.

As shown in Figure \ref{fig:late-interaction}, while the encoder uses the mention's context (determined by the maximum
number of tokens that the backbone model can process) to produce contextualized embeddings, only the mention tokens are
retained in $\boldsymbol{Q}$. Afterwards, XTR computes a token-level similarity matrix $\boldsymbol{P} \in
	\mathbb{R}^{m \times n}$, which is defined by the pairwise similarities between each mention token and each entity
token: \mbox{$\boldsymbol{P}_{ij} = \boldsymbol{Q}_{i}^\top\boldsymbol{D}_{j}$}.

To determine which of these token-level scores contribute to the final mention-entity similarity, XTR uses a binary
alignment matrix $\boldsymbol{A} \in \{0,1\}^{n \times m}$. This matrix, has two formulations depending on whether the
score is being is computed at test time ($\boldsymbol{A}$) or during training ($\boldsymbol{\hat{A}}$). At test time
XTR uses $\boldsymbol{A}_{ij} = \mathds{1}_{[j = \operatorname{argmax}_{j'} \boldsymbol{P}_{ij}]}$. This is the
\enquote{sum-of-max} operator as originally introduced in ColBERT, which yields the following scoring function:

\begin{equation}\label{eq:xtr:lt}
	f(\boldsymbol{Q}, \boldsymbol{D}) = \frac{1}{n} \sum_{i=1}^{n} \sum_{j=1}^{m} \boldsymbol{A}_{ij} \boldsymbol{P}_{ij} = \frac{1}{n} \sum_{i=1}^{n} \max_{1 \le j \le m} \boldsymbol{Q}_{i}^{\top} \boldsymbol{D}_{j}
\end{equation}

Intuitively, the final alignment score is determined by each query token's best match. XTR departs from ColBERT in how
it defines the alignment matrix during training. That is, when optimizing model parameters XTR uses instead
$\boldsymbol{\hat{A}}_{ij} = \mathds{1}_{[j \in \operatorname{top-k}(\boldsymbol{P}_{ij'})]}$, where
$\operatorname{top-k}$ operator spans all tokens within a mini-batch of size $B$ (i.e., $1 \le j' \le mB$). The
intuition behind this choice is that the $\operatorname{top-k}$ operator simulates the inference stage, where only the
entities whose tokens are retrieved by the $\operatorname{top-k}$ operation are considered for the final ranking. This
forces the model \textit{during training} to consider only token similarities high enough to be globally retrieved,
leading to stronger training signal\footnote{See the original study for details.} \citep{lee2023rethinking}.


\subsubsection{Name-based vs entity-based}%
A KB can be represented in two ways: (a) by entity or (b) by names. In the first case, a candidate entity $e$ is
represented by concatenating all of its known names. In the second case, each name is treated as a separate candidate.

We explore both representations. Given a set of candidates $\mathcal{C} = \{ \boldsymbol{D}^{1}, \cdots,
	\boldsymbol{D}^{k} \}$ (either entities or names), we define the probability of a candidate $\boldsymbol{D}^{i}$ being
the correct link for a given mention $\boldsymbol{Q}$ (under the model parameters $\theta$) as follows:

\begin{equation}\label{eq:xtr:prob}
	p(\boldsymbol{D}^{i} \mid \boldsymbol{Q}; \theta) = \frac{\exp\left(f(\boldsymbol{Q}, \boldsymbol{D}^{i})\right)}{\sum_{j=1}^{\vert \mathcal{C} \vert} \exp\left(f(\boldsymbol{Q}, \boldsymbol{D}^{j})\right)}
\end{equation}

We train two separate models based on these representations. For a given mention $\boldsymbol{Q}$, the entity-based
model optimizes the standard cross-entropy loss \citep{bridle1990probabilistic}. For the name-based model, we optimize
instead the maximum marginal likelihood (MML) objective proposed by \citep{biomedicalentisung2020}. Formally, for a
single mention $\boldsymbol{Q}$ the loss is defined as follows:

\begin{equation}\label{eq:xtr:mml}
	\mathcal{L}_{\text{MML}} = -\log \sum_{i=1}^{|\mathcal{C}|} \mathds{1}_{[V_C(\boldsymbol{Q}) = V_{\text{KB}}(\boldsymbol{D}^{i})]} \, p(\boldsymbol{D}^{i} \mid \boldsymbol{Q})
\end{equation}

where $\mathds{1}_{[*]}$ is an indicator function, $V_C : \boldsymbol{Q} \rightarrow e$ returns the gold KB entity
associated with a mention $\boldsymbol{Q}$, and similarly $V_KB : \boldsymbol{D} \rightarrow e$ returns the entity
associated the entity name $\boldsymbol{D}^{i}$. Intuitively, this objective encourages the embeddings of
$\boldsymbol{Q}$ and all $\mathcal{C} = \{ \boldsymbol{D}^{1}, \cdots, \boldsymbol{D}^{k} \}$ that are associated to
the same entity to be close in the embedding space.

For the name-based model we preprocess the KB with the \textit{homonym disambiguation} (HD) approach introduced by
\citet{belhd}. The preprocessing is necessary to handle homonyms (names shared by multiple concepts) which otherwise
prevent unique linking predictions. For instance, HD expands the name \enquote{conorenal syndrome} into
\enquote{conorenal syndrome (short rib-polydactyly syndrome)} and \enquote{conorenal syndrome (Mainzer-Saldino
	disease)} for \mesh:D012779 and \mesh:C535463, respectively. For \gene entities,  we follow
\cite{10.1093/bioinformatics/btag011} and include species information to every gene name (e.g., \enquote{A2M (alpha2-microglobulin, human)}).

\subsubsection{Temperature scaling}%
Cosine similarity is the de facto standard similarity metric in late-interaction models (including XTR)
\citep{DBLP:conf/cikm/ChaffinS25}. For models optimized via contrastive losses, scaling the cosine similarities has a
significant impact on downstream performance\footnote{ Since their values are bounded within $[-1, 1]$, their gradients
	can easily vanish during softmax normalization \citep{wang2021understanding}. } \citep{A_simple_framew_Chen_2020}.
However, neither ColBERT nor XTR explicitly mention taking this into account. Therefore, during training, we scale all scores obtained from
Eq. \ref{eq:xtr:lt} by a learnable temperature parameter $\tau$ as defined in \cite{Learning_transf_Radfor_2021}.

\subsubsection{Query expansion}\label{sec:xtr:method:model:qe}%
Mentions often lack the information required for disambiguation. For instance, a disease may have a dominant or
recessive form, yet a specific mention might only use the general name.
If present, this information can be found in the surrounding context. Therefore, a model should somehow identify and
\enquote{encode} it in the mention's token embeddings.

Importantly, this information is explicitly contained within the target entities. For instance, as shown in Figure
\ref{fig:late-interaction}, for every \gene entity we always include its associated species. As multi-vector models are
based on token-level similarities, we hypothesize that expanding the mention tokens to account for this information
will improve downstream performance.

When introducing ColBERT, \cite{Colbert_Effici_Khatta_2020} argued that a BERT-based model can leverage \texttt{[MASK]}
tokens to perform query expansion\footnote{Since a BERT-based model learns to predict plausible missing tokens for each
	\texttt{[MASK]} during pre-training \citep{bertpretraindevlin2019}, the intuition is that during retrieval training,
	the model will naturally learn to expand these tokens into terms missing from the query that aid candidate retrieval.}
(QE). However, they did not introduce an explicit training signal to encourage this behavior. In fact,
\cite{giacalone2024beneath} found that \texttt{[MASK]} embeddings tend to cluster close to existing query tokens, thus
performing term weighting rather than true expansion.

To address this limitation, \belxtr introduces an auxiliary objective that explicitly 
encourages \texttt{[MASK]} embeddings to represent information missing from the mention.
Unlike ColBERT, and unlike XTR, which does not use \texttt{[MASK]} tokens, we supervise QE using the target entity $\boldsymbol{D}^{+}$.
Specifically, for each mention, we use a trigram-based similarity model to select the
gold entity name most similar to the mention. By filtering out tokens already present in the mention, we isolate
a set of target expansion tokens: $\mathcal{Y} = \{ \boldsymbol{d}^{+}_{1}, \dots, \boldsymbol{d}^{+}_{n} \}$.
The number of expansion tokens determines the number of \texttt{[MASK]} tokens
to be appended to each contextualized mention.
At inference time, the gold entity is unavailable.
We therefore estimate the required number of \texttt{[MASK]} tokens using the same filtering heuristic,
but applied to the most similar entity name retrieved from the entire KB,
rather than from the set of gold entity names.

%
QE is formulated as a
multiple-instance learning problem \citep{dietterich1997solving}. Formally, for a single \texttt{[MASK]} embedding
$\boldsymbol{q}_{i}$ the loss is defined as:

\begin{equation}
	\mathcal{L}_{\text{QE}} = -\log \frac{\exp\left(\max_{\boldsymbol{d}_j \in \mathcal{Y}} \boldsymbol{q}_{i}^{\top}\boldsymbol{d}_j\right)}{\sum_{\boldsymbol{d}_k \in \mathcal{D}^{+}} \exp\left(\boldsymbol{q}_{i}^{\top}\boldsymbol{d}_k\right)}
\end{equation}

The loss encourages each \texttt{[MASK]} embedding to assign its highest similarity 
within the positive entity to one missing expansion token,
rather than to a token already present in the mention.
\footnote{For instance,
	given the mention \{
	\enquote{$\alpha2$} , \enquote{micro} , \enquote{globulin}, \enquote{\texttt{[MASK]}} \}, and the candidate \{
	\enquote{$\alpha2$} , \enquote{micro} , \enquote{globulin}, \enquote{rat} \}, the objective requires the similarity
	between \enquote{\texttt{[MASK]}} and \enquote{rat} to be higher than the similarity between \enquote{\texttt{[MASK]}}
	and any other token.
}
Without an additional constraint,
multiple \texttt{[MASK]} embeddings may collapse onto the same expansion token.
We therefore introduce a dispersion objective \citep{Contrastive_Lea_Wang_2024}
that encourages \texttt{[MASK]} embeddings to be dissimilar among each other:
$
\mathcal{L}_{\mathrm{DIS}}
=
\frac{1}{\vert \mathcal{Y} \vert}
\sum_{i < j}
\boldsymbol{q}_i^{\top}\boldsymbol{q}_j
$

The model is trained via multi-task learning \citep{Multitask_Learn_Caruan_1997} with the final objective function for
a single mention being $\mathcal{L} = \mathcal{L}_{\text{MML}} + \lambda(\mathcal{L}_{\text{QE}} +
	\mathcal{L}_{\text{DIS}})$, where $\lambda$ is a hyperparameter (see Appendix \ref{app:training} for details).

\subsection{Evaluation protocol}\label{sec:xtr:method:eval}

We evaluate \belxtr across three experimental settings to provide a comprehensive assessment of its performance. First,
we compare it against state-of-the-art neural methods (Section \ref{sec:xtr:method:eval:nn}). Second, we assess its
effectiveness as a candidate generator for LLM-based reranking (Section \ref{sec:xtr:method:eval:llm}). Finally, we
compare it with traditional (non-neural) type-specific systems (Section \ref{sec:xtr:method:eval:tts}), which remain
the established standard for production-level BEL \citep{10.1093/jamiaopen/ooae129,Integrating_AI_Wieger_2025}.

\subsubsection{Neural retrievers}\label{sec:xtr:method:eval:nn}%

\indent \textbf{Task} Our comparison with neural approaches adopts the in-KB formulation of BEL (no NIL label)
\citep{gerbilbenchmroder2018} and evaluates models on human-annotated (gold) mentions. Performance is reported using
micro-averaged recall@1 (accuracy).

\textbf{Data} We utilize the corpora and KBs provided by the BELB benchmark \citep{belb}
(See Appendix \ref{app:belb} for an overview of corpora and Ks.)
We evaluate on ten corpora linked to six different KBs\footnote{We exclude corpora of the \variant entity type due to
	the scalability limitations of current neural systems.}. For \ncbigene, we use the subsets determined by the species of
the genes in the \gnormplus and \nlmgene corpora (see Appendix \ref{app:gene_subsets}). This reflects a common
real-world use case where often only a specific subset of species is relevant for linking \citep{wei2012}.

\textbf{Methods} We compare our model against the following state-of-the-art
neural approaches: BioSyn \citep{biomedicalentisung2020}, GenBioEL \citep{generativebiomyuan2022}, BELHD \citep{belhd}
ANGEL \citep{kim-etal-2025-learning-negative}, arboEL \citep{agarwal2022}, and KRISSBERT \citep{zhang2022knowledge}.
We include in the comparison as well the GenBioEL variant with HD (GenBioEL+HD), as \cite{belhd}
show it produces significantly better results.
With the exception of KRISSBERT and ANGEL, all models are trained from scratch on BELB (see Section \ref{sec:xtr:discussion:nn}).

For KRISSBERT, we report the results directly from the original study (without second-stage reranking). We do this
because the authors only provide code and data for the \enquote{supervised} variant\footnote{See:
	\url{https://huggingface.co/microsoft/BiomedNLP-KRISSBERT-PubMed-UMLS-EL/tree/main/usage}.}, which can solely link to
entities present in the training data. For ANGEL, we obtain predictions exclusively for the BELB corpora for which the
authors released trained model checkpoints. This restriction applies because training ANGEL models across all BELB
corpora is computationally prohibitive, and the originally reported results rely on a lenient evaluation\footnote{See:
	\url{https://github.com/dmis-lab/ANGEL/blob/main/train_positive_only.py}.}, which overestimates model performance
\citep{zhang2022knowledge}.

\subsubsection{LLM-based reranking}\label{sec:xtr:method:eval:llm}%

\indent \textbf{Task} For LLM-based candidate reranking
we replicate the experimental setting proposed by \cite{10.1093/bioinformatics/btag011}.
In this setting, the entity linking model is used as a candidate generator to retrieve the $k$ highest-scoring entities for each mention from the KB.
The mention, its context, and the $k$ candidates are passed to an LLM, which is tasked with selecting the correct entity among the candidates.
Performance of the entire pipeline is reported using micro-averaged recall@1 (accuracy).

\textbf{Data} We use the corpora and KBs provided by BELB for the following entity types:
\gene, \species, \disease, and \chemical.
An important difference from the setting reported in Section \ref{sec:xtr:method:eval:nn}
is that the results are obtained on the \textit{refined} test sets of the corpora \citep{tutubalina2020}.
These sets exclude duplicate test mentions or those that overlap with mentions in the training data,
providing a significantly harder testbed.
Secondly, mentions linked to multiple concepts in the KB (composite mentions) are excluded from the evaluation.
This is the setup proposed by \cite{10.1093/bioinformatics/btag011}, which we follow to allow direct comparison.

\textbf{Methods} We compare our model against BioSyn+GRF, the retriever proposed by \cite{10.1093/bioinformatics/btag011}. Given a mention and its
context, BioSyn+GRF prompts OpenAI's GPT-4o \citep{OpenAI_2024_GPT4o} to generate various types of
mention-specific information, such as a context-aware definition or a list of synonyms (an approach known as Generative Relevance Feedback (GRF) \citep{mackie2023generative}).
A fine-tuned BioSyn model \citep{biomedicalentisung2020} is then used to obtain embeddings of both the mention and the LLM-generated
feedback, which are combined into a single representation to retrieve candidates from the KB.

For the reranking stage, we evaluate both our model and BioSyn+GRF using the strategy proposed by
\cite{10.1093/bioinformatics/btag011}. Specifically, we employ GPT-4o as the reranker and use their system
prompt\footnote{ See: \url{https://github.com/dash-ka/Biomedical-Entity-Linking-GRF/blob/master/select_candidate.py}.},
which provides the model with a task-specific instruction, $k=10$ candidates and the sentence containing the mention as
context.

We report as well the performance of BeLink \citep{BeLink_Biomedi_Shlyk_2026}, a retrieve-and-rerank pipeline using
SapBERT \citep{SelfAlignmentLiuF2021} as retriever, and a 8B Qwen3 model \citep{Qwen3_Technical_Yang_2025}
instruction-tuned as reranker for the BEL task. The BeLink's retriever uses as well an LLM to enrich the mentions
before retrieval, but with a Qwen3-14B model. As the authors do not release the instruction-tuned model we cannot
compare the effect of using \belxtr with this reranker.

\subsubsection{Traditional type-specific methods}\label{sec:xtr:method:eval:tts}%

\indent \textbf{Task} To compare against traditional type-specific methods, we replicate the experimental setting of PubTator3 \citep{pubtator3},
which evaluates models on the end-to-end document-level entity linking task.
Specifically, for each document, a third-party NER model is used to identify entity mentions,
which are subsequently resolved by a linking model.
The set of unique entities is then treated as document-level classes.
The performance of the complete pipeline is reported using micro-averaged precision, recall, and F1-score.

\textbf{Data} All results are obtained on the \biored corpus \citep{bioredarichluol2022},
which offers linking annotations for the same entity types as BELB.
The inputs to the linking models are mentions identified by the AIONER model  \citep{aioner}.
The target KBs are determined by the identified entity type (e.g., \ctddiseases for \disease).
With the exception of the \chemical entity type\footnote{PubTator3 uses \mesh instead of \ctdchemicals.},
all evaluated methods use the same KBs as BELB, although not necessarily the exact same version (see Section \ref{sec:xtr:discussion:tts}).

\textbf{Methods} We compare against the traditional type-specific systems integrated into PubTator3.
We train \belxtr models on the same corpora used to train the corresponding type-specific system in PubTator3
(see Appendix \ref{app:pubtator3} for details on PubTator3's models and corpora).

\section{Results}\label{sec:xtr:results}

We present the results of our empirical validations. First we report the ablation study used to determine the best
\belxtr configuration (Section \ref{sec:xtr:results:ablation}). We then report the results when comparing against: (i)
state-of-the-art neural models (Section \ref{sec:xtr:results:neural}), (ii) LLM-powered methods (Section
\ref{sec:xtr:results:rerank}) and (iii) traditional type-specific systems (Section \ref{sec:xtr:results:trad}).

\subsection{Ablation study}\label{sec:xtr:results:ablation}

\begin{table*}[!htbp]
	\centering
	\begin{tabular}{l|l|l|l}
		\toprule
		                                    & \textbf{\ctddiseases}     & \textbf{\ctdchemicals}   & \textbf{\ncbigene}              \\
		                                    & (\textbf{\disease})       & (\textbf{\chemical)}     & (\textbf{\gene})                \\
		\midrule
		                                    & \textbf{\ncbidisease}     & \textbf{\bcfivecdr}      & \textbf{\nlmgene}               \\
		\midrule
		\belxtr (name-based)                & \textbf{90.34}            & \textbf{96.22}           & 90.66                           \\
		\quad 1) entity-based               & $85.42_{\downarrow4.92}$  & $93.44_{\downarrow2.78}$ & $69.44_{\downarrow21.22}$       \\
		\quad 2) no temp. scaling           & $77.39_{\downarrow12.95}$ & $86.76_{\downarrow9.45}$ & $75.08_{\downarrow15.57}$       \\
		\quad 3.1) add \texttt{[MASK]}      & $89.45_{\downarrow0.89}$  & $95.21_{\downarrow1.01}$ & $90.03_{\downarrow0.63}$        \\
		\quad 3.2) add \texttt{[MASK]} + QE & $89.96_{\downarrow0.48}$  & $96.04_{\downarrow0.17}$ & $\mathbf{91.24}_{\uparrow0.58}$ \\
		\bottomrule
	\end{tabular}
	\caption{
		Ablation study of modifications of XTR \citep{lee2023rethinking} introduced in \belxtr (see Section \ref{sec:xtr:method:model}).
		Performance is mention-level recall@1 on the \textit{development set} of the corpora.
		\textbf{Bold} indicates best score. $\uparrow$/$\downarrow$ indicates increase/decrease in performance w.r.t. the baseline.
		QE: active training of query expansion (see Section \ref{sec:xtr:method:model:qe}).
	}\label{tab:xtr_ablations}
\end{table*}

\textbf{1) Entity- vs name-based} In Table \ref{tab:xtr_ablations}, we observe that a name-based KB yields superior performance for a multi-vector
model. This can be explained by the mechanics of token-level alignments.
The XTR variant of Eq. \ref{eq:xtr:lt}
requires the model to maximize the similarity of all mention tokens with \textit{all} tokens of an entity\footnote{See Appendix
	A in \citep{lee2023rethinking}}. If an entity contains multiple names with distinct surface forms, an entity-based
representation may destabilize training, by forcing the model to maximize similarities across unrelated tokens.

\textbf{2) Temperature scaling} Additionally, we see that training without temperature scaling in the loss computation causes a drop of $\sim$9 to
$\sim$15 percentage points in recall@1 across corpora. This confirms the importance of this optimization technique
found in previous studies \citep{A_simple_framew_Chen_2020,wang2021understanding,Learning_transf_Radfor_2021} as well
for entity linking.

\textbf{3) Query expansion} As mentioned in Section \ref{sec:xtr:method:model}, XTR does not make use of \texttt{[MASK]} tokens.
Therefore, to evaluate our QE strategy we trained \belxtr with two different settings.
One reflects the baseline (3.1), where, like ColBERT,
we add \texttt{[MASK]} tokens to each mention but optimize only the retrieval loss (Eq. \ref{eq:xtr:mml}).
The other one (3.2) instead explicitly trains
the \texttt{[MASK]} embeddings to perform QE (see Section \ref{sec:xtr:method:model:qe}).

For \ncbidisease and \ctdchemicals, we observe that the inclusion of \texttt{[MASK]} tokens degrades downstream
performance (3.1), even when our custom objective loss is used (3.2). We attribute this drop to the inconsistent
expansion requirements across mentions. For instance, expansion tokens in \disease mentions vary from different
spellings (e.g., expanding \enquote{tumor} with \enquote{tumour}) to multiple tokens which may not appear in the
mention's context (e.g., expanding \enquote{attenuated polyposis} with \enquote{familial}, \enquote{adenomatous}, and
\enquote{coli}).

This hypothesis is further supported by our results on \nlmgene, where active QE (3.2) yields a performance benefit,
albeit marginal. \gene mentions require short and highly consistent expansions, primarily in the form of species names.
For all subsequent experiments, we train \belxtr models with active QE only on entity types that inherently require
species information (\gene and \cellline) for disambiguation.

\subsection{Neural retrievers}\label{sec:xtr:results:neural}

\begin{table*}[!h]
	\centering
	\resizebox{\textwidth}{!}{
		\begin{tabular}{l|ll|ll|l|ll|ll|l}
			\toprule
			                                & \multicolumn{2}{c|}{\textbf{\ctddiseases}} & \multicolumn{2}{c|}{\textbf{\ctdchemicals}} & \multicolumn{1}{c|}{\textbf{\cellosaurus}} & \multicolumn{2}{c|}{\textbf{\ncbigene}} & \multicolumn{2}{c|}{\textbf{\ncbitaxonomy}} & \multicolumn{1}{c}{\textbf{\umls}}                                                                                         \\
			                                & \multicolumn{2}{c|}{(\textbf{\disease})}   & \multicolumn{2}{c|}{(\textbf{\chemical)}}   & \multicolumn{1}{c|}{(\textbf{\cellline})}  & \multicolumn{2}{c|}{(\textbf{\gene})}   & \multicolumn{2}{c|}{(\textbf{\species})}    &                                                                                                                            \\
			\midrule
			                                & \textbf{\ncbidisease}                      & \textbf{\bcfivecdr}                         & \textbf{\bcfivecdr}                        & \textbf{\nlmchem}                       & \textbf{\bioid}                             & \textbf{\gnormplus}                & \textbf{\nlmgene} & \textbf{\esseight} & \textbf{\linnaeus} & \textbf{\medmentionspv} \\
			\midrule
			\textit{Name-based}             &                                            &                                             &                                            &                                         &                                             &                                    &                   &                    &                    &                         \\
			\quad BioSyn                    & 79.90                                      & 84.83                                       & 84.57                                      & 70.35                                   & 80.79                                       & OOM                                & OOM               & 82.79              & \textbf{88.60}     & OOM                     \\
			\quad GenBioEL                  & 82.71                                      & 88.29                                       & 94.60                                      & 75.00                                   & 94.79                                       & 6.80                               & 2.89              & 88.27              & 76.92              & 41.16                   \\
			\quad GenBioEL+HD               & 83.02                                      & 88.20                                       & 94.15                                      & 74.10                                   & \underline{96.30}                           & 66.08                              & \underline{66.43} & \textbf{89.96}     & 77.62              & 64.59                   \\
			\quad BELHD                     & \underline{87.60}                          & \underline{89.23}                           & 92.93                                      & \textbf{82.39}                          & \textbf{96.99}                              & \underline{77.84}                  & 59.03             & 84.35              & 81.89              & \textbf{70.58}          \\
			\quad ANGEL                     & 84.06                                      & 88.77                                       & 94.51                                      & -                                       & -                                           & -                                  & -                 & -                  & -                  & 58.22                   \\
      \quad \belxtr (ours)            & \textbf{88.75}                             & \textbf{89.87}                              & \textbf{95.46}                             & \underline{80.56}                       & 95.72                                       & \textbf{84.82}                     & \textbf{82.29}    & \underline{88.14}  & \underline{82.03}  & \underline{69.47}       \\
			\textit{Entity-based}           &                                            &                                             &                                            &                                         &                                             &                                    &                   &                    &                    &                         \\
			\quad arboEL$\dagger$           & 80.00                                      & 84.87                                       & 87.40                                      & 71.76                                   & 95.02                                       & 34.64                              & 29.96             & 78.62              & 74.97              & 68.67                   \\
			\quad KRISSBERT$\dagger$$\ddag$ & 82.80                                      & 85.0                                        & \underline{95.10}                          & -                                       & -                                           & -                                  & -                 & -                  & -                  & 61.30                   \\
			\bottomrule
		\end{tabular}
	}
	\caption{Performance (mention-level recall@1) of neural models on the \textit{test} set of BELB corpora.
		\textbf{Bold} and \underline{underlined} indicate best and second best score, respectively.
    HD: Homonym Disambiguation \citep{belhd}. OOM: out-of-memory ($>$200GB) $\dagger$ Without cross-encoder
		reranking
    $\ddag$ Results reported in \citep{zhang2022knowledge} (see Section \ref{sec:xtr:method:eval:nn}).
	}~\label{tab:results:xtr}
\end{table*}

In Table \ref{tab:results:xtr}, we observe that \belxtr outperforms all baseline models on five out of the ten corpora,
while ranking second on the remaining ones (except for \bioid). Performance gains vary across entity types. For
\disease and \chemical, the improvement over the previous state-of-the-art is marginal. However, we emphasize that on
the \chemical subset of \bcfivecdr, \belxtr outperforms methods that underwent task-specific pre-training, whether on a
large-scale corpus (KRISSBERT) or directly on the target KB (ANGEL).

The most substantial improvements are achieved on the \gene corpora. We argue that \belxtr's competitive advantage here
stems primarily from two factors. First, unlike existing embeddings methods like BELHD, \belxtr utilizes the maximum
context window allowed by its backbone model (see Section \ref{sec:xtr:method:model}). This enables it to capture
broader contextual information, which is crucial for gene normalization since species information is not always located
in close proximity to the gene mention \citep{assigningspeciluol2022}.

The second critical factor is \belxtr's ability to model fine-grained token similarities (see Section
\ref{sec:xtr:dicussion:gene} for detailed discussion). In contrast, standard embedding methods must compress these
nuanced distinctions into a single vector, making it harder to resolve such granular differences. Generative methods
like GenBioEL can capture fine-grained information as well \citep[a]{decao2021autoregressive}. However, their
optimization is limited the generation of a single output, while \belxtr can directly optimize for the ranking of the
candidate set, explaining its advantage \citep[b]{highlyparalleldecao2021}.


\subsection{LLM-based reranking}\label{sec:xtr:results:rerank}

\begin{table*}[!h]
	\centering
	\resizebox{0.8\textwidth}{!}{
		\begin{tabular}{l|ll|ll|ll|ll}
			\toprule
			                          & \multicolumn{2}{c|}{\textbf{\ctddiseases}} & \multicolumn{2}{c|}{\textbf{\ctdchemicals}} & \multicolumn{2}{c|}{\textbf{\ncbigene}} & \multicolumn{2}{c}{\textbf{\ncbitaxonomy}}                                                                                     \\
			                          & \multicolumn{2}{c|}{(\textbf{\disease})}   & \multicolumn{2}{c|}{(\textbf{\chemical)}}   & \multicolumn{2}{c|}{(\textbf{\gene})}   & \multicolumn{2}{c}{(\textbf{\species})}                                                                                        \\
			                          & \textbf{\ncbidisease}                      & \textbf{\bcfivecdr}                         & \textbf{\bcfivecdr}                     & \textbf{\nlmchem}                          & \textbf{\gnormplus} & \textbf{\nlmgene} & \textbf{\esseight} & \textbf{\linnaeus} \\
			\midrule
			\multicolumn{9}{c}{Retrieve}                                                                                                                                                                                                                                                                    \\
			\midrule
			BioSyn$\dagger$           & 71.35                                      & 74.69                                       & 82.01                                   & \underline{70.90}                          & 69.15               & 33.80             & 61.61              & \underline{69.06}  \\
			BioSyn+GRF$\dagger$       & \textbf{74.75}                             & \textbf{76.69}                              & \textbf{94.00}                          & \textbf{77.58}                             & \underline{78.68}   & \underline{39.12} & \underline{76.05}  & \textbf{83.42}     \\
			\belxtr (ours)            & \underline{73.89}                          & \underline{76.12}                           & \underline{89.04}                       & 70.18                                      & \textbf{81.72}      & \textbf{78.65}    & \textbf{85.92}     & 59.57              \\
			\midrule
			\multicolumn{9}{c}{Retrieve-and-Rerank}                                                                                                                                                                                                                                                         \\
			\midrule
			BioSyn+GRF+GPT4o$\dagger$ & \textbf{76.21}                             & \textbf{79.32}                              & \underline{92.71}                       & \textbf{82.58}                             & \underline{85.26}   & \underline{69.17} & \underline{75.00}  & \textbf{91.16}     \\
			\belxtr (ours)+GPT4o      & \underline{75.37}                          & \underline{78.77}                           & 91.23                                   & \underline{80.23}                          & \textbf{87.21}      & \textbf{75.96}    & \textbf{81.69}     & 70.21              \\
			BeLink-4B$\ddag$          & 72.9                                       & 77.0                                        & \textbf{93.5}                           & 76.7                                       & 80.4                & 47.6              & 74.2               & 87.2               \\
			BeLink-8B$\ddag$          & 73.4                                       & 76.6                                        & \textbf{93.5}                           & 77.4                                       & 81.5                & 51.6              & 73.9               & \underline{90.0}   \\
			\bottomrule
		\end{tabular}
	}
	\caption{
		Comparison of \belxtr with BioSyn and BioSyn with Generative Relevance Feedback (BioSyn+GRF) \citep{10.1093/bioinformatics/btag011}
		as standalone retrievers and in a LLM-based retrieve-and-rerank pipeline.
		Performance is recall@1 on the \textit{refined test set} of the corpora \citep{tutubalina2020} (see Section \ref{sec:xtr:method:eval:llm}).
		\textbf{Bold} and \underline{underlined} indicate best and second best score, respectively.
		$\dagger$ Results reported in \citep{10.1093/bioinformatics/btag011}
		$\ddag$ Results reported in \citep{BeLink_Biomedi_Shlyk_2026}
	}
	\label{tab:xtr:llm}
\end{table*}




\textbf{Retrieve} In Table \ref{tab:xtr:llm},
we see that \belxtr achieves performance comparable to the LLM-enhanced BioSyn+GRF.
The largest gap between the standalone models occurs on \chemical corpora.
We attribute this to the fact that BioSyn+GFR uses an LLM to generate the standard scientific name for the given mention
(among other expansions) to be used by the BioSyn model.
As \chemical entities present a high naming variability \citep{krallinger2015chemdner},
leveraging the LLM's domain-specific knowledge provides an advantage over models like \belxtr,
which instead rely on labeled data to learn different entity names .

\textbf{Retrieve-and-rerank} When utilized as candidate generator for LLM-based reranking,
\belxtr is outperformed by BioSyn+GRF on \disease and \chemical corpora.
We attribute this to optimization trade-offs: \belxtr optimizes for high-precision retrieval of the correct candidate,
whereas BioSyn+GRF is specifically designed to increase recall,
making its candidate sets inherently better suited for subsequent reranking.

Nevertheless, \belxtr maintains a definitive advantage on \gene corpora. This is especially evident on \nlmgene, where
standalone \belxtr not only outperforms BioSyn+GRF, but remarkably, also beats the BioSyn+GRF+GPT-4o pipeline (see
Section \ref{sec:xtr:discussion:llm} for discussion). The performance degradation when combining \belxtr with GPT-4o on
this corpus can be explained by the restricted context window used during LLM reranking (i.e., only the sentence
containing the mention). In multiple cases, the sentence does not contain explicit species information, causing GPT-4o
to either fail to generate a response or select candidates associated with the most common species (e.g., human or
mouse).

\subsection{Traditional type-specific methods}\label{sec:xtr:results:trad}

\begin{table*}[!htbp]
	\centering
	\resizebox{\textwidth}{!}{
		\begin{tabular}{l|lll|lll|lll|lll|lll}
			\toprule
			                   & \multicolumn{3}{c|}{\textbf{\disease}} & \multicolumn{3}{c|}{\textbf{\chemical}} & \multicolumn{3}{c|}{\textbf{\cellline}} & \multicolumn{3}{c|}{\textbf{\gene}} & \multicolumn{3}{c}{\textbf{\species}}                                                                                                                                                                                                         \\
			\midrule
			                   & P                                      & R                                       & F1                                      & P                                   & R                                     & F1                & P                 & R                 & F1                & P                 & R                 & F1                & P                 & R                 & F1                \\
			\midrule
			PubTator3$\dagger$ & 75.33                                  & 83.43                                   & 79.17                                   & \textbf{83.26}                      & 80.63                                 & \underline{81.92} & 76.00             & 86.36             & 80.85             & \textbf{90.60}    & 79.41             & \textbf{84.63}    & 93.97             & 96.46             & 95.20             \\
			BELHD              & 80.34                                  & \textbf{88.72}                          & \textbf{84.32}                          & \underline{79.50}                   & \textbf{88.89}                        & \textbf{83.93}    & \textbf{83.97}    & \textbf{90.38}    & \textbf{87.06}    & 76.20             & \textbf{85.32}    & 80.50             & \underline{95.47} & \textbf{97.64}    & \textbf{96.55}    \\
			\belxtr            & \textbf{80.62}                         & \underline{86.25}                       & \underline{83.34}                       & 78.26                               & \underline{85.01}                     & 81.50             & \underline{82.69} & \underline{86.54} & \underline{84.57} & \underline{82.91} & \underline{84.84} & \underline{83.86} & \textbf{95.65}    & \underline{97.28} & \underline{96.46} \\
			\bottomrule
		\end{tabular}
	}

	\caption{Comparison with PubTator3 \citep{pubtator3} and BELHD \citep{belhd} on document-level end-to-end entity linking.
		\textbf{Bold} and \underline{underlined} indicate best and second best score, respectively.
		Results are obtained from mentions in the \textit{test set} of \biored \citep{bioredarichluol2022} identified by AIONER \citep{aioner}.
		P: Precision, R: Recall, F1: F1 score.
		$\dagger$ Results reported in \citep{pubtator3}.
	}
	\label{tab:xtr:vs_pubtator}
\end{table*}

In Table \ref{tab:xtr:vs_pubtator}, we compare \belxtr against the type-specific systems integrated into PubTator3 on
document-level entity linking. The results show that \belxtr outperforms PubTator3's models on \disease, \cellline and
\species, while obtains comparable performance on \chemical and \gene. The latter is particularly noteworthy, as
PubTator3 relies on GNorm2 for gene disambiguation, a highly specialized and manually tuned system
\citep{gnorm2animprweic2023}.

When compared with the previous best neural model (BELHD), we observe that, although \belxtr achieves higher
performance on the in-corpus evaluation for \disease and \chemical (see Table \ref{tab:results:xtr}), its performance
on \biored, though comparable, is lower than that of BELHD. This reflects a well-documented phenomenon in the
literature: neural models are susceptible to distribution shifts, and a performance increase on a single corpus does
not necessarily translate to better performance across different distributions
\citep{galea2018exploiting,TowardsReliablGiorgi2020, hunflair2}.


%


\section{Discussion}\label{sec:xtr:discussion}

We introduced \belxtr, a multi-vector model for biomedical entity linking that delivers state-of-the-art performance.
Our method and evaluation protocols are subject to specific design choices and limitations, which we discuss in more
detail below.

\subsection{Neural retrievers}\label{sec:xtr:discussion:nn}

We show that \belxtr outperforms six state-of-the-art neural approaches for BEL on five out of ten corpora in the
standardized BELB benchmark, while ranking second on four (see Section~\ref{sec:xtr:results:neural}). We emphasize
that, due to high computational costs, we did not perform extensive hyperparameter tuning for the competing methods
(e.g., searching for the optimal learning rate) but instead relied on the settings reported by the original authors.

Consequently, it is possible that these methods could achieve higher performance if fully tuned. Nevertheless, we
believe this setup provides the fairest possible comparison across methods. BEL studies on first-stage retrievers
exhibit stark differences in preprocessing, corpora, and experimental setups, which makes direct comparisons based on
published numbers problematic \citep{zhang2022knowledge,acomprehensivekartch2023,belb}.

\subsection{LLM-based reranking}\label{sec:xtr:discussion:llm}

Our experiments show that, on gene disambiguation \belxtr alone achieves a higher recall@1 than a LLM-based
retrieve-and-rerank pipeline.

However, the reranking setup proposed by \cite{10.1093/bioinformatics/btag011} (which we follow) has three important
aspects to consider when interpreting the results. First, the LLM only utilizes the sentence in which the mention
appears. This is particularly relevant for \gene corpora, as the target species necessary for disambiguation is
frequently absent from the immediate local context \citep{assigningspeciluol2022}. Second, their prompt does not make
use of enhancements like few-shot examples or chain-of-thoughts.\footnote{Although
	\cite{10.1093/bioinformatics/btag011} note that their simple prompt delivered better performance than the
	Chain-of-Thought (CoT) prompting proposed by \citet{dobbins2025generalizable}.} Third, the chosen LLM (GPT-4o) is no
longer the state-of-the-art. It is therefore possible that (a) integrating a wider context, (b) optimizing the prompt
for the task (e.g., with DSPy \citep{khattab2024dspy}), or (c) employing more recent LLMs could yield superior results.

We argue that \belxtr's performance without reranking on \gene mentions is particularly appealing for a core real-world
application of entity linking: information extraction pipelines over large document collections like PubMed
\citep{10.1093/jamiaopen/ooae129,Integrating_AI_Wieger_2025}. In such deployments, computational efficiency is
paramount. While \cite{10.1093/bioinformatics/btag011} demonstrate that local, small-scale LLMs (e.g., 1B to 7B
parameters) offer competitive results, maximizing performance on challenging entity types like \gene still requires
proprietary LLMs. Running these models at scale remains prohibitively costly and inefficient
\citep{OpportunitiesATian2023}.

\subsection{Traditional type-specific methods}\label{sec:xtr:discussion:tts}

Our results show that \belxtr is highly competitive with state-of-the-art, type-specific systems. Most notably, on the
challenging \gene mentions, its performance closely approaches that of the highly specialized GNorm2
\citep{gnorm2animprweic2023}, which relies on manually curated mappings to resolve normalization edge cases.

Although both \belxtr and PubTator3 target the same underlying KBs, they utilize slightly different versions, which may
affect comparability \citep{belb}. However, we expect the impact of this discrepancy to be minimal, as both systems
employ KB versions released within a similar timeframe\footnote{2023--2024.}. The only exception occurs with the
\chemical entity type, where PubTator3 maps to \mesh while \belxtr targets \ctdchemicals. In this case, \belxtr holds a
slight advantage. As \ctdchemicals is a subset of \mesh, the resulting smaller candidate space inherently simplifies
the linking task \citep{Unified_Examina_Ong_N_2024}.


\subsection{Gene disambiguation}\label{sec:xtr:dicussion:gene}

\begin{figure}[htbp]
	\centering
	\begin{minipage}[b]{0.48\textwidth}
		\centering
		\includegraphics[width=\textwidth]{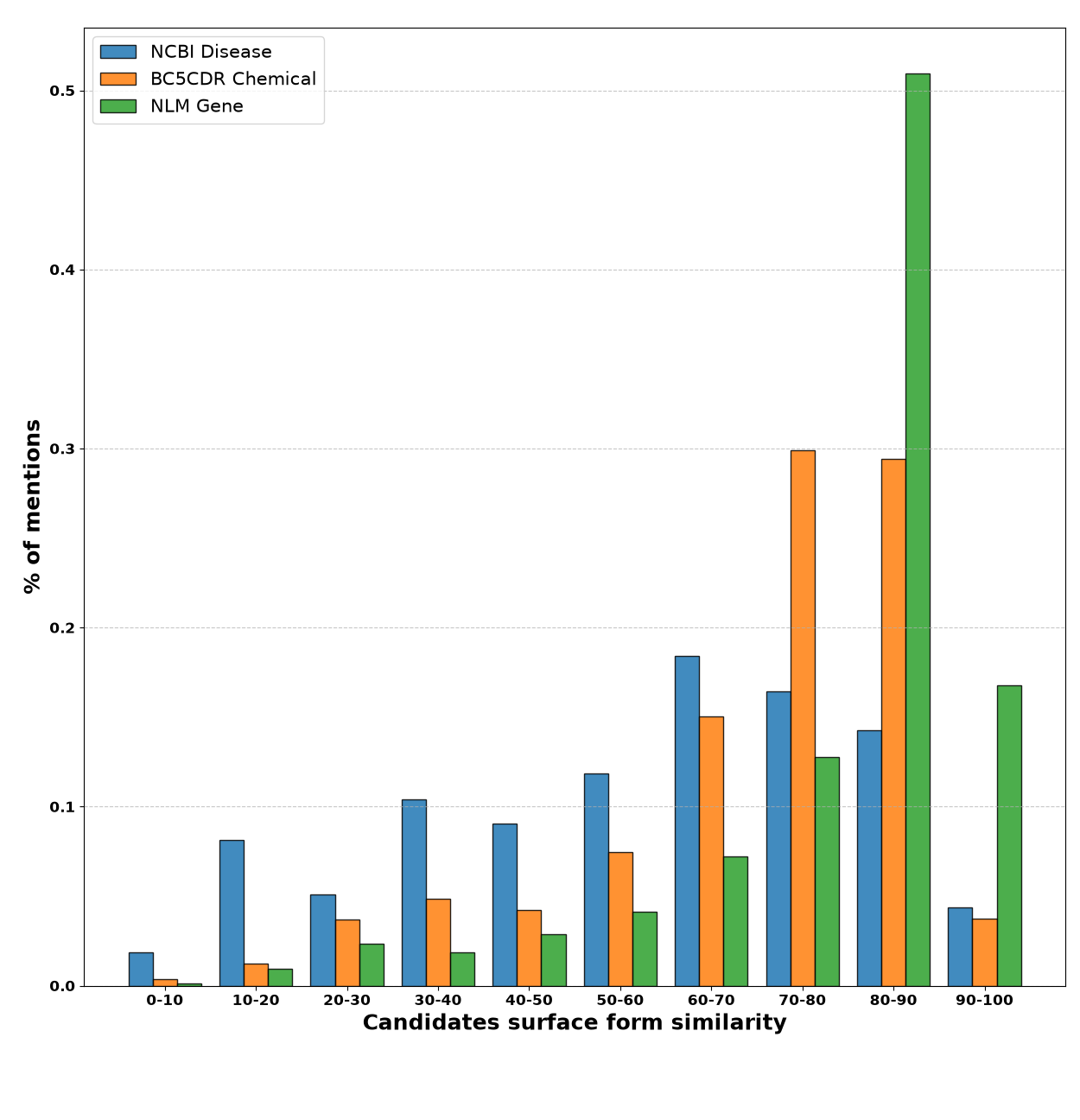}
		\caption{Number of mentions stratified by name similarity in the candidate set in a \disease (\ncbidisease), \chemical (\bcfivecdr) and \gene (\nlmgene) corpus.
		}
		\label{fig:name_density}
	\end{minipage}

	\vspace{0.5cm} 

	\begin{minipage}[b]{0.48\textwidth}
		\centering
		\includegraphics[width=\textwidth]{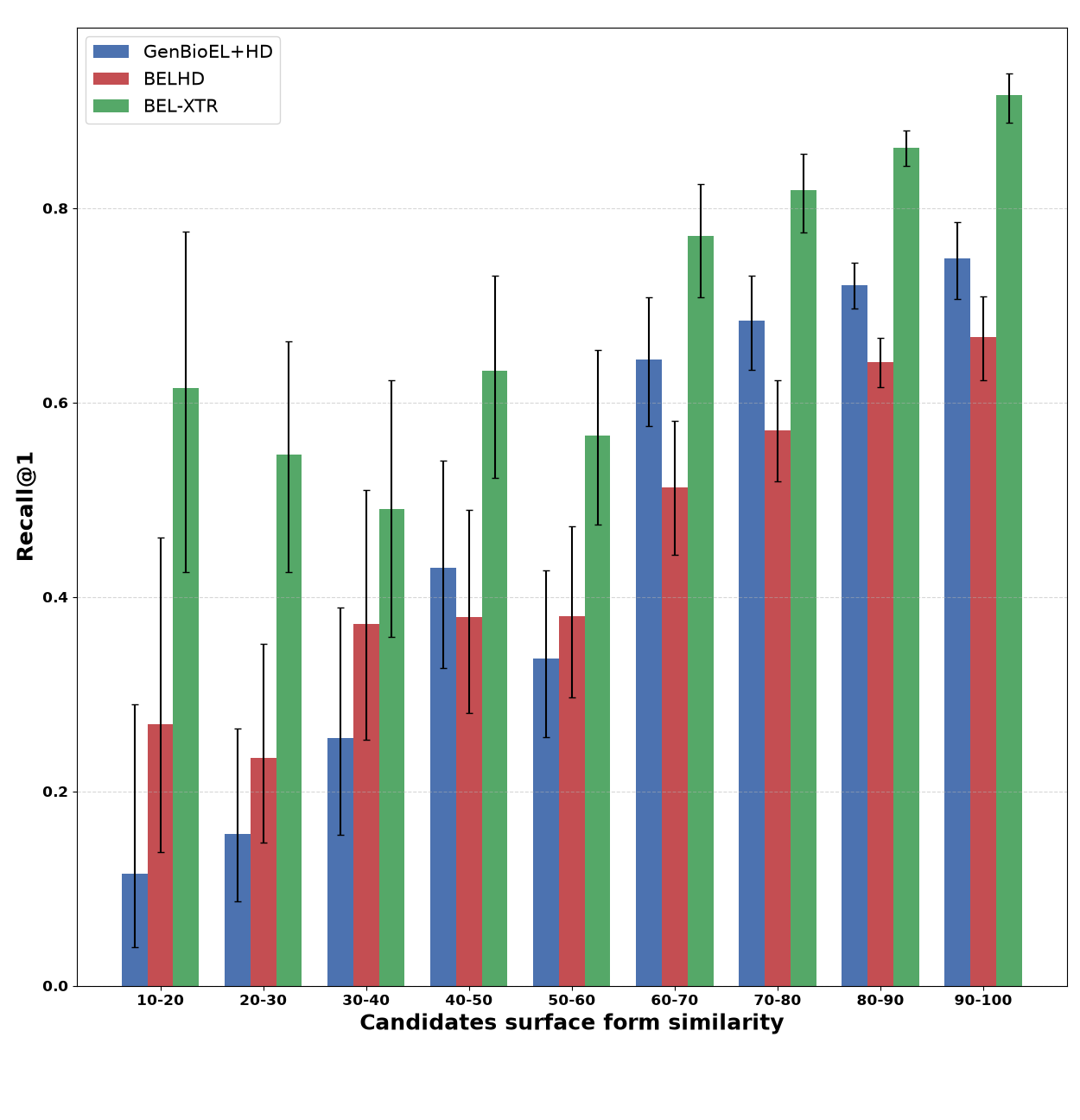}
		\caption{Recall@1 of GenBioEL \citep{generativebiomyuan2022} with HD (see Section \ref{sec:xtr:method:eval:nn}),
			BELHD \citep{belhd} and \belxtr (ours)
			on the \nlmgene corpus stratified by name similarity in the candidate set.
		}
		\label{fig:gene_linking}
	\end{minipage}
\end{figure}


In our experiments, \belxtr yields the largest performance gains on gene disambiguation. We argue that this advantage
stems from its design, which allows it to capture fine-grained name variations.

To illustrate this, we perform the following analysis on three corpora representative of different entity types. Using
a trigram-based similarity model, we retrieved (a) the gold entity name and (b) the top-5 false positive names most
similar to each test mention. We then categorized each mention by the average Levenshtein distance
\citep{marzal1993computationon} between its gold name and top-5 false positives. This distance score reflects candidate
set similarity: a score $\le10$ means false positives are easily distinguishable via string matching, whereas a score
$\ge90$ indicates near-identical candidates.

As shown in Figure \ref{fig:name_density}, $\sim$70\% of mentions in the gene corpus (\nlmgene) have candidate
similarity scores $\ge80$. Consequently, even when a model correctly identifies the species and retrieves top candidate
names, separating the gold entity from false positives requires detecting subtle differences. For example, 33 human
genes belong to the \enquote{ADAM} family and differ by only one or two characters. Chemical mentions also exhibit this
pattern (e.g., \enquote{Lithium-3} vs. \enquote{Lithium-4}), though many fall in the 70-80 similarity range where
differences are more structurally pronounced (e.g., \enquote{6-oxymelatonin} vs. \enquote{7-azamelatonin}).

Figure \ref{fig:gene_linking} shows that \belxtr achieves its peak recall@1 with the tightest confidence intervals on
high-similarity mentions, explaining its edge over rival name-based models. While autoregressive methods can capture
fine-grained patterns \citep{decao2021autoregressive}, multi-vector architectures perform better here because they
directly optimize candidate set rankings. In contrast, generative models like GenBioEL train on single-name targets
\citep{highlyparalleldecao2021}. Autoregressive models like ANGEL \citep{kim-etal-2025-learning-negative} which use
negative samples during training can potentially bridge this performance gap.

\subsection{Latency and memory footprint}\label{sec:xtr:dicussion:efficiency}

\begin{table}[!htbp]
	\centering
	\resizebox{0.5\textwidth}{!}{
		\begin{tabular}{l|l|l|l}
			\toprule
			                           & \textbf{Recall@1} & \textbf{Abstract/s} & \textbf{Index (GB)} \\
			\midrule
			Single-vector (name-based) & 57.93             & 15.94 ($\pm$0.11)   & 0.75                \\
			Multi-vector (name-based)  & 82.29             & 2.38 ($\pm$0.07)    & 5.90                \\
			\bottomrule
		\end{tabular}
	}
	\caption{
		Comparison of performance (recall@1), inference speed (average abstract/second) and memory footprint (index size) between
		a  single- and multi-vector bi-encoder. Results are obtained processing the test split of \nlmgene
		on a NVIDIA GeForce RTX 3090 (three independent runs).
	}
	\label{tab:xtr:efficiency}
\end{table}

As shown in Table~\ref{tab:xtr:efficiency}, when compared to single-vector models, multi-vector retrievers, although
providing superior performance, come at the cost of: (a) increased inference latency and (b) a larger memory footprint.
This inherently limits accessibility and ease of deployment.

The latency issue can be mitigated by the use of approximate nearest neighbor (ANN) search methods, such as HNSW
\citep{malkov2018efficient}, though this trade-off typically reduces recall. Additionally, with the rising popularity
of late-interaction models, specialized retrieval engines have been developed that substantially improve inference
speed while fully preserving retrieval quality \citep{santhanam2022plaid, WARP_An_Effici_Scheer_2025}.

W.r.t. the memory footprint issue, numerous index compression techniques are available for nearest neighbor search
(e.g., product quantization \citep{jegou2010product}) which can significantly reduce memory overhead. Additionally,
many entity names are highly redundant, differing only by minor orthographic variations (e.g., \enquote{tumor} vs.
\enquote{tumour}). As contextualized token retrieval is less sensitive to such lexical variations than standard lexical
matching, these names can be safely subsampled. For instance, a clustering algorithm could be applied each the names of
each entity to retain only the variant closest to a cluster center. We leave the exploration of these optimization
techniques to future work.


\subsection{Conclusion}\label{sec:xtr:conclusion}

We presented \belxtr, an multi-vector model for biomedical entity linking designed to exploit granular subword
similarities between mentions and entity names. \belxtr improves upon XTR \citep{lee2023rethinking} by specifically
tailoring the model to biomedical KBs. Through extensive experiments we show that \belxtr sets a new state-of-the-art
on five out of ten standard evaluation corpora. On the challenging cross-species gene disambiguation \belxtr achieves a
higher recall@1 than an LLM-powered retrieve-and-rerank pipeline and closely matches a specialized rule-based method.
Our results highlight multi-vector models as a practical alternative to hard-to-maintain rule-based system or scenarios
where LLM-based reranking is too costly as in PubMed-scale mining.

%
%
%
%

\bibliographystyle{abbrvnat}
\bibliography{reference}

\begin{appendices}

	\section{Biomedical Entity Linking Benchmark}\label{app:belb}

\begin{table*}[!h]
	\centering
	\resizebox{\textwidth}{!}{
		\begin{tabular}{l|c|c|l|l|l|l}
      \toprule
			                                                            & \textbf{Version}     & \textbf{History} & \textbf{Entities} & \textbf{Names} & \textbf{Synonyms} & \textbf{Homonyms (PN)}  \\
			  \midrule                                                                                                                                                             %
			\disease                                           &                      &                  &                   &                &                   &                         \\
			\quad \ctddiseases \citep{comparativetoxdavis2023}  & monthly $\dagger$    & \cross           & 13,188            & 88,548         & 6.71              & 0.39\% (-)              \\
			  \hline                                                                                                                                                               %
			\chemical                                          &                      &                  &                   &                &                   &                         \\
			\quad \ctdchemicals \citep{comparativetoxdavis2023} & monthly  $\dagger$   & \cross           & 175,663           & 451,410        & 2.56              & - (-)                   \\
			  \hline                                                                                                                                                               %
			\cellline                                          &                      &                  &                   &                &                   &                         \\
			\quad \cellosaurus \citep{thecellosaurusbairoc2018} & -                    & \tick           & 144,568           & 251,747        & 1.74              & 3.21\% (1.22\%)         \\
			  \hline                                                                                                                                                               %
			\species                                           &                      &                  &                   &                &                   &                         \\
			\quad \ncbitaxonomy \citep{thencbitaxonoscott2012}  & -                    & \tick           & 2,491,364         & 3,783,882      & 1.51              & 0.04\% (-)              \\
			  \hline                                                                                                                                                               %
			\gene                                              &                      &                  &                   &                &                   &                         \\
			\quad \ncbigene \citep{brown2015}                   & -                    & \tick           & 42,252,923        & 105,570,090    & 2.49              & 47.37\% (8.32\%)        \\
			\qquad \gnormplus subset                                    &                      &                  & 703,858           & 2,455,772      & 3.48              & 50.79\% (9.13\%)        \\
			\qquad \nlmgene subset                                      &                      &                  & 873,015           & 2,913,456      & 3.33              & 53.61\% (9.55\%)        \\
			  \hline                                                                                                                                                               %
			\variant                                           &                      &                  &                   &                &                   &                         \\
			\quad \dbsnp \citep{sherry2001}                     & build 156  $\dagger$ & \tick           & 1,053,854,063     & 3,119,027,235  & 2.95              & 1,557,105,418 (49.92\%) \\
			  \midrule                                                                                                                                                             %
			\quad \umls \citep{bodenreider2004}                 & 2017AA (full)        & -                & 3,464,809         & 7,938,833      & 2.29              & 2.07\% (0.16\%)         \\
      \bottomrule
		\end{tabular}
	}
	\caption{Overview of the KBs available in BELB according to their entity type.
		We report the number of entities, synonyms per entity, homonyms and how many of them are the primary name (PN).
		$\dagger$ No archive of previous versions is provided
	}~\label{tab:kbs:overview}
\end{table*}

\begin{table*}[!h]
	\centering
	\resizebox{\textwidth}{!}{
		\begin{tabular}{l|c|c|c|c}
      \toprule
			                                                               & \textbf{Documents (train / dev / test)} & \textbf{Annotations (train / dev / test)} & \textbf{0-shot entity} & \textbf{0-shot name} \\
                                                                     \midrule
			\disease                                              &                                         &                                           &                        &                      \\
			\quad \ncbidisease \citep{dogan2014}                   & 592 / 100 / 100                         & 5,133 / 787 / 960                         & 150 (15.62\%)          & 185 (19.27\%)        \\
      \quad \bcfivecdr \citep{li2016a}          & 500 / 500 / 500                         & 4,149 / 4,228 / 4,363                     & 388 (8.89\%)           & 765 (17.53\%)        \\
			\hline
			\chemical                                             &                                         &                                           &                        &                      \\
			\quad \bcfivecdr \citep{li2016a}           & 500 / 500 / 500                         & 5,148 / 5,298 / 5334                      & 1,038 (19.46\%)        & 415 (7.78\%)         \\
			\quad \nlmchem $\dagger$ \citep{nlmchembc7mislama2022} & 80 / 20 / 50                            & 20,796 / 5,234 / 11514                    & 3,908 (33.94\%)        & 1,534 (13.32\%)      \\
			\hline
			\cellline                                             &                                         &                                           &                        &                      \\
			\quad \bioid $\ddag$ \citep{arighi2017bio}             & 231 / 59 / 60                           & 3,815 / 1,096 / 864                       & 158 (18.29\%)          & 45 (5.21\%)          \\
			\hline
			\species                                              &                                         &                                           &                        &                      \\
			\quad \linnaeus $\dagger$ \citep{linnaeusaspemartin}   & 47 / 17 / 31                            & 2,115 / 705 / 1,430                       & 385 (26.92\%)          & 58 (4.06\%)          \\
			\quad \esseight \citep{pafilis2013}                    & 437 / 63 / 125                          & 2,557 / 384 / 767                         & 363 (47.33\%)          & 107 (13.95\%)        \\
			\hline
			\gene                                                 &                                         &                                           &                        &                      \\
			\quad \gnormplus \citep{wei2015}                       & 279 / 137 / 254                         & 3,015 / 1,203 / 3,222                     & 2,822 (87.59\%)        & 163 (5.06\%)         \\
			\quad \nlmgene \citep{islamaj2021}                     & 400 / 50 / 100                          & 11,263 / 1,371 / 2,729                    & 1,215 (44.52\%)        & 353 (12.94\%)        \\
			\hline
			\variant                                              &                                         &                                           &                        &                      \\
			\quad \snp \citep{challengesintthomas2011}             & - / - / 292                             & - / - / 517                               & -                      & -                    \\
			\quad \osiris \citep{osirisv12anfurlon2008}            & - / - / 57                              & - / - / 261                               & -                      & -                    \\
			\quad \tmvarthree \citep{wei2022}                      & - / - / 214                             & - / - / 1,018                             & -                      & -                    \\
			\midrule
			\umls                                                 &                                         &                                           &                        &                      \\
			\quad \medmentionspv \citep{mohanmedmentions}          & 2,635 / 878 / 879                       & 122,178 / 40,864 / 40,143                 & 8,167 (20.34\%)        & 7,945 (19.79\%)      \\
      \bottomrule
		\end{tabular}
	}
	\caption{Overview of the corpora available in BELB with their primary characteristics: number of documents, annotations
		and how many of them are zero-shot by entity or by name.
		$\ddag$ Full text
		$\ddag$ Figure captions}~\label{tab:corpora:overview}
\end{table*}

In Table \ref{tab:kbs:overview} and \ref{tab:corpora:overview} we provide an overview of the KBs and corpora available in the BELB benchmark,
respectively.

\section{NCBI Gene subsets}\label{app:gene_subsets}

\begin{table*}[!htbp]
	\centering
	\begin{tabular}{l|l|l}
		\toprule
		\multicolumn{2}{c|}{\textsc{NCBI Taxonomy}} &                                                                 \\
		\midrule
		Entity                                      & Name                                      & Corpora             \\
		\midrule
		3055                                        & Chlamydomonas reinhardtii                 & \nlmgene            \\
		3702                                        & thale cress                               & \gnormplus,\nlmgene \\
		3847                                        & soybean                                   & \gnormplus          \\
		4896                                        & fission yeast                             & \gnormplus,\nlmgene \\
		6239                                        & Caenorhabditis elegans                    & \gnormplus,\nlmgene \\
		6956                                        & European house dust mite                  & \nlmgene            \\
		7227                                        & fruit fly <Drosophila melanogaster>       & \gnormplus,\nlmgene \\
		7955                                        & zebrafish                                 & \gnormplus,\nlmgene \\
		8355                                        & African clawed frog                       & \gnormplus,\nlmgene \\
		8364                                        & tropical clawed frog                      & \gnormplus,\nlmgene \\
		9031                                        & chicken                                   & \gnormplus,\nlmgene \\
		9606                                        & human                                     & \gnormplus,\nlmgene \\
		9615                                        & dog                                       & \nlmgene            \\
		9823                                        & pig                                       & \gnormplus,\nlmgene \\
		9913                                        & cattle                                    & \gnormplus,\nlmgene \\
		9940                                        & sheep                                     & \nlmgene            \\
		9986                                        & rabbit                                    & \gnormplus,\nlmgene \\
		10029                                       & Chinese hamster                           & \nlmgene            \\
		10089                                       & Ryukyu mouse                              & \nlmgene            \\
		10090                                       & house mouse                               & \gnormplus,\nlmgene \\
		10116                                       & Norway rat                                & \gnormplus,\nlmgene \\
		10298                                       & Herpes simplex virus type 1               & \gnormplus          \\
		11676                                       & Human immunodeficiency virus 1            & \gnormplus,\nlmgene \\
		11709                                       & Human immunodeficiency virus 2            & \nlmgene            \\
		11908                                       & Human T-cell leukemia virus type I        & \gnormplus          \\
		41856                                       & Hepatitis C virus genotype 1              & \gnormplus          \\
		51031                                       & New World hookworm                        & \nlmgene            \\
		81972                                       & Arabidopsis lyrata subsp. lyrata          & \nlmgene            \\
		333760                                      & Human papillomavirus type 16              & \gnormplus          \\
		511145                                      & Escherichia coli str. K-12 substr. MG1655 & \gnormplus          \\
		559292                                      & Saccharomyces cerevisiae S288C            & \gnormplus,\nlmgene \\
		2886926                                     & Escherichia phage P1                      & \nlmgene            \\
		\bottomrule
	\end{tabular}
	\caption{\ncbigene subsets determined by the species (\ncbitaxonomy entities) of the gene mentions in \gnormplus and \nlmgene}\label{tab:ncbi_gene_subsets}
\end{table*}

In Table \ref{tab:ncbi_gene_subsets} we report the \ncbigene subsets determined by the species (\ncbitaxonomy entities)
of the gene mentions in \gnormplus and \nlmgene.

\section{PubTator3}\label{app:pubtator3}

\begin{table*}[!h]
	\centering
	\resizebox{\textwidth}{!}{
		\begin{tabular}{l|l|l|l}
			\toprule
			          & \textbf{Entity Linking}                     & \textbf{Train}                                           & \textbf{KB}                                  \\
			\midrule
			\disease  & TaggerOne $\dagger$ \citep{leaman2016}      & \ncbidisease and \bcfivecdr (train and validation split) & CTD Diseases \citep{comparativetoxdavis2023} \\
			\chemical & NLM-Chem \citep{nlm_chem_a_new_islama_2021} & \nlmchem (full corpus)                                   & MeSH \citep{lipscomb2000medical}             \\
			\cellline & TaggerOne $\dagger$ \citep{leaman2016}      & \bioid (full corpus)                                     & Cellosaurus \citep{thecellosaurusbairoc2018} \\
			\gene     & GNorm2 \citep{gnorm2animprweic2023}         & \gnormplus and \nlmgene (train and validation split)     & NCBI Gene \citep{brown2015}                  \\
			\species  & SpeAss \citep{assigningspeciluol2022}       & -                                                        & NCBI Taxonomy \citep{thencbitaxonoscott2012} \\
			\bottomrule
		\end{tabular}
	}
	\caption{Overview of the entity linking models used by PubTator3. $\dagger$ Improved version introduced by \cite{pubtator3}}\label{tab:pubtator3}
\end{table*}

PubTator3 \citep{pubtator3} is a biomedical literature resource which aggregates state-of-the-art tools for biomedical
information extraction. In Table \ref{tab:pubtator3} we provide an overview of all entity linking models deployed in
PubTator3. We note that the entity linking (normalization) model are non-neural type-specific methods. The KBs used
for normalization are gathered from inspecting the original publication of each tool.

When comparing with PubTator3 we train \belxtr models on the corpora reported in Table \ref{tab:pubtator3}. As common in
cross-corpus evaluations the full corpus is used for training \citep{galea2018exploiting, hunflair2}, unless it
overlaps with \biored, which is sampled from \nlmgene, \ncbidisease, \bcfivecdr and \tmvarthree. For training a
\species model we rely on the combination of \linnaeus and \esseight (full corpora) as they have no overlap with
\biored.

\section{Training details}\label{app:training}

Each \belxtr model is trained for a maximum of five epochs with a mini-batch size of 4 queries (mentions with context).
We use negative mining to collected 32 hard negatives for each query,
and every query uses the all negatives in the mini-batch as negatives: this totals to  a maximum of 128 negatives per query.
For setting the hyperparameters we use the development set of the \ncbidisease corpus, which are then used for all experiments.
The learning rate is set to $3e-6$. We explore the following values for the $\operatorname{top-k}$: \{ 128, 256, 512, 1024, 2048 \}
and find 2048 to work the best for both training and inference.
The $\lambda$ hyperparameter for the loss function is set to $0.3$.

\end{appendices}

\end{document}